\documentclass[letterpaper, 10 pt, conference]{ieeeconf}  

\IEEEoverridecommandlockouts
\IEEEoverridecommandlockouts                              
\usepackage{xcolor}
\usepackage{soul}
\usepackage{cite}
\usepackage{amsfonts}
\usepackage{graphicx}
\usepackage{textcomp}
\usepackage{xcolor}
\def\BibTeX{{\rm B\kern-.05em{\sc i\kern-.025em b}\kern-.08em
    T\kern-.1667em\lower.7ex\hbox{E}\kern-.125emX}}
\usepackage{subcaption}
\usepackage[colorlinks]{hyperref}
\usepackage{amssymb}
\usepackage[ruled,longend,algo2e]{algorithm2e}
\usepackage{amsmath}
\usepackage{algorithm}
\usepackage{multirow}
\usepackage{booktabs}
\usepackage{caption}
\usepackage[utf8]{inputenc}
\usepackage{tabularray}
\UseTblrLibrary{booktabs}

\DeclareMathOperator*{\argmax}{arg\,max}

\graphicspath{{figures}}
\begin{document}

\title{{Consensus-based Resource Scheduling for Collaborative Multi-Robot Tasks}}

\author{Nazish Tahir \and Ramviyas Parasuraman
\thanks{The authors are with the Heterogeneous Robotics Research Lab (HeRoLab), School of Computing, University of Georgia, Athens, GA 30602, USA. Authors email: {\tt\small \{nazish.tahir,ramviyas\}@uga.edu}}}

\maketitle

\begin{abstract}

We propose integrating the edge-computing paradigm into the multi-robot collaborative scheduling to maximize resource utilization for complex collaborative tasks, which many robots must perform together. Examples include collaborative map-merging to produce a live global map during exploration instead of traditional approaches that schedule tasks on centralized cloud-based systems to facilitate computing. Our decentralized approach to a  consensus-based scheduling strategy benefits a multi-robot-edge collaboration system by adapting to dynamic computation needs and communication-changing statistics as the system tries to optimize resources while maintaining overall performance objectives. Before collaborative task offloading, continuous device, and network profiling are performed at the computing resources, and the distributed scheduling scheme then selects the resource with maximum utility derived using a utility maximization approach. Thorough evaluations with and without edge servers on simulation and real-world multi-robot systems demonstrate that a lower task latency, a large throughput gain, and better frame rate processing may be achieved compared to the conventional edge-based systems.

\end{abstract}

\section{Introduction}
Due to a recent shift towards developing intelligent Cyber-Physical Systems (CPS), utilizing the cloud to accelerate robotic computing tasks is not new. Although a robot may have sufficient computing resources to perform small-scale operations, it relies heavily on remote resources to perform large-scale computations. Offloading enables robot operators to access data from anywhere at any time by providing global storage as well as processing\cite{spatharakis2022resource, tahir2022analog}. Since the concept of CPS heavily relies on the timely delivery of suitable data placement to the relevant computing entities \cite{uysal2021semantic}, robots that move resource-intensive tasks to remote resources require efficient solutions with low latency.

The idea of cloud computing has been extended to multi-robot systems as cloud robotics \cite{7482658}. According to this paradigm, the cloud offers computing resources like virtual machines or containers and resources from nearby and far-off data centers, enabling scalable and massive data processing. However, this objective presents its own set of design challenges, including data processing, offloading decision-making, resource allocation, and controller architecture.

An intelligent resource allocation strategy holds significance in resource-constrained computational systems since it aims to mitigate resource contention while ensuring performance guarantees for the effective utilization of cloud and edge systems \cite{tahir2023mobile,penmetcha2021deep}. As a result, this article formulates the problem of finding the optimal computational resource for offloading robotic computational-intensive tasks, a topic of extensive research \cite{wan2016cloud,hu2012cloud}, especially if the task is collaborative in nature and requires real-time processing through multiple resources. However, cloud robotics has its fundamental disadvantage for a persistent connection to an external network. It requires traversing the local network, leading to excessive delays caused due to network congestion or poor link quality.

Utilizing network edges allows sequential activities to be carried out simultaneously, prevents bigger files from being offloaded, and does away with data pre-processing, all of which can reduce the total latency of multi-robot collaborative tasks \cite{9279239}. Additionally, it is less expensive than a full unload to the cloud service. Thus, the offloading decisions and resource allocation in order to achieve the trade-off between computational costs and performance efficiency are critical research problems.

\begin{figure*}[t]
    \centering
    \includegraphics[width=0.99\textwidth]{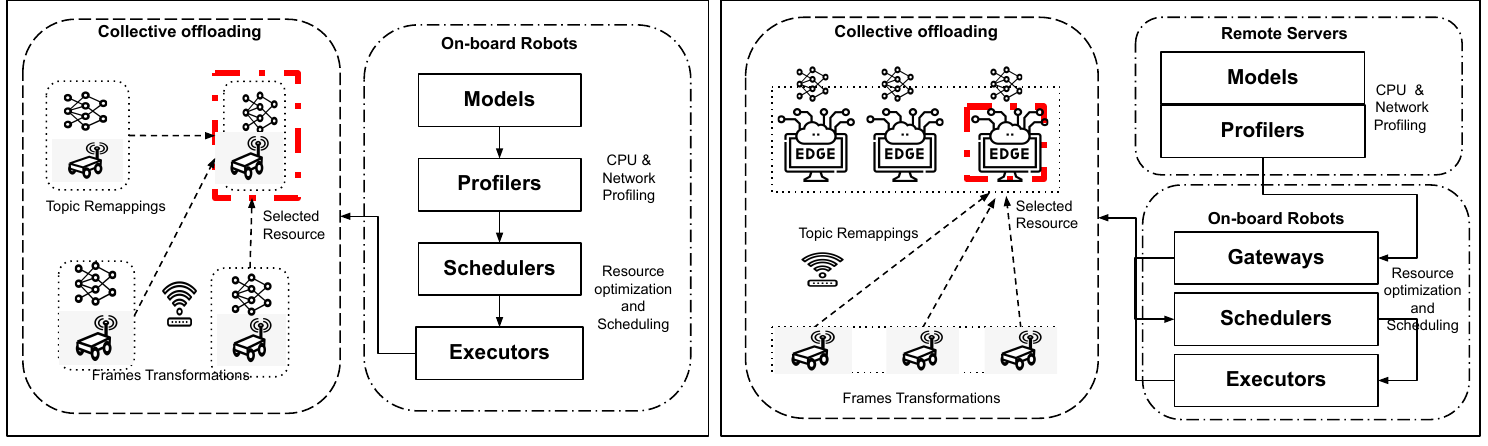}
    \caption{Distributed utility-aware collaborative scheduling for robot-robot and multi-robot-edge based systems}
    \label{fig:edge_robotics}
\end{figure*}

This paper proposes allocating compute-intensive multi-robot collaborative tasks that require offloading to an optimal computational resource (edge server or robot). While finding optimal resources, the proposed strategy minimizes performance costs during offloading, and overall task latency is reduced compared to static deployment of edge servers. The thematic overview of the proposed solution is shown in Fig.~\ref{fig:edge_robotics} with robot-robot and robot-edge collaborative scenarios. The proposed method is tested against conventional static offloading methods in a multi-robot collaborative scenario requiring remote resources, specifically multi-robot collaborative computing, for offloading real-time map-merging to see if the goals of optimal resource utilization and task latency minimization have been met.

We make the following contributions in this paper. 
\begin{itemize}
  \item We propose a novel consensus-based collaborative scheduling scheme for robots to optimize resources, enabling robots to perform resource-hungry tasks beyond their onboard capabilities.
  \item The proposed strategy utilizes a utility maximization-based decision-making process, adjusting offloading based on processing power, memory, and network quality, aiming to minimize latency and enhance overall performance.
  \item With the design of distributed schedulers and executors, we calculate utility and set weights according to the dynamics of the system and network conditions to alleviate the computational strain from the resource-constrained platform and enhance performance. 
  \item We evaluate the efficacy of our proposed approach with a collaborative multi-robot map-merging task on computing resources, edge servers, or robots. 
  \item The effectiveness of the suggested multi-robot-edge system is assessed in terms of map accuracy, throughput, structural similarity index, map frame frequency, overall execution time, and resource optimization (CPU and memory monitoring). A comparison is made between (i) Fixed deployment (tasks are pre-initialized on edges regardless of the change in system dynamics) and (ii) Dynamic deployment (proposed utility-based collective task offloading mechanism which adapts to current running states and offloads to one computing resource).
  
\end{itemize}

\section{Related Work}
Existing literature is rife with scheduling with resource optimization objectives \cite{dechouniotis2022edge,huang2021efficient,8279382}. Direct offloading full tasks to the Cloud server introduces network congestion, high task latency, more probability of a single point of failure, and reduced battery life. 

Many seminal works in tackling the resource constraints posed by local onboard computing in executing SLAM algorithms have turned to edge computing to improve efficiency and reduce task latency. P. Huang et al. present a collaborative multi-robot laser SLAM ColaSLAM \cite{Huang2021} that leverages the edge computing concept for SLAM execution optimization. The robot-edge synergy uses edge computing to enable robots to execute SLAM and produce a global map. 
Researchers of Spatharakis et al. \cite{spatharakis2022resource} provide a set-based estimation for robot offloading mechanism in the context of edge robotics. An offloading strategy is proposed to compensate for the uncertainty of the local estimation techniques with more accurate remotes while finding the balance between navigation accuracy and mission duration. The offloading framework keeps the network conditions and computation availability in mind, together with maintaining the stability of the control system. 

Solving the resource optimization problem on robots, authors in Ben Ali et al. \cite{ben2022edge} propose the implementation of ORB-SLAM2 using edge computing to offload parts of Visual-SLAM. They split the architecture of the algorithm between the edge and the robot by keeping the tracking module on the robot while moving more intensive computational modules, i.e., local mapping and loop closure to the edge. They hope to lower computational and memory overhead by connecting the components tightly without compromising performance precision. 
Another recent work \cite{9723464} presents a reputation-based collaborative robotic learning framework (CoRoL) for computational offloading in collaborative robots with the ability to isolate the impact of malicious or poor-performing robots on computational task execution. 

A more recent work \cite{zhu2024collor} focuses on collaborative computational offloading for robots based on a multi-criteria utility function that also considers the robots' energy efficiency and formulates a joint offloading and routing problem. Another paper \cite{9978606} in the realm of robot perception tried to address the safety challenges posed by occlusions in autonomous driving by proposing a multi-tier perception task offloading framework by the collaborative computing approach using autonomous vehicles and roadside units (RSUs) aiming to reduce processing delay due to computational offloading. 

This paper departs from the above-mentioned works by proposing a cooperative multi-robot scheduling scheme that aims to find the optimal computational resource for collaborative task offloading and maximize the overall system performance. The proposed scheme considers various factors such as task computational complexity, communication overhead, and individual robot/edge capabilities to allocate tasks efficiently among the computational server.

\section{System Dynamics and Profiling} 
\subsection{System Overview}
We propose a collaborative scheduling strategy for offloading tasks within multi-robot or multi-edge-based systems, where robots can execute their collaborative task within themselves or to the edge servers connected via wireless communication channels in close proximity for faster task execution. This is different from typical task allocation problems in the literature, where N tasks need to be allocated to M resources (usually $M \geq N$) \cite{saeik2021task}. 

In our case, the problem is to find the optimal computational resource (an edge or a robot in a multi-robot-edge system) for the group of robots, which needs to perform a collaborative task as a group. The task should require all robots' contributions, and all robots would benefit from this collaborative task in their decisions or planning \cite{yang2019self}. An example of such a task is multi-robot map merging, where all the robots perform individual mapping of their environment and, at the same time, offload their mapping data to a remote resource for feature extraction to merge these multiple maps together as one global map which is downloaded by the robots to be used for navigation. 

\subsection{Multi-robot System} 
The multi-robot system comprises a group of ground or aerial robots that communicate using the Robot Operating System (ROS) (https://www.ros
.org/) Communication framework. To ensure communication between the robots and the edges, ROS services (Request/Reply communication) or topics (Publish/Subscribe communication) are employed. A node on any computing entity can publish messages on a specific topic, whereas all other nodes that subscribe to this topic will receive the message. The ROS Master adds name registration and lookup functionality to the computation graph and allows nodes to communicate with one another, exchange messages, and activate services. 

\subsection{Edge Servers} To provide computational offloading service to numerous robots concurrently, we deploy edge servers adjacent to the group of MRS for reduced job latency than typical cloud servers. Following receipt of an offloaded job from a robot, the server will conduct the task on the client's behalf and, upon completion, will deliver the output result to the robot. Due to robots statically offloading some compute-intensive tasks to the edge servers, those servers may be overloaded or operating at maximum capacity. The continuous resource profilers quantify the computational resources made available by each edge server to be shared among the robots. The machine learning inference, classification, or fusion modules would also operate on the edge servers in the background for the coordinated task execution.

\subsection{Resource Profiling}
Three profilers are continually active on the computing devices (either edges or robots), monitoring system parameters and delivering them to the individual resource gateways through ROS topics. The profilers are crucial in identifying computational and communication bottlenecks in the overall multi-robot system. 

\subsubsection{Device Profiler} represents the operating condition of the edge device and measures CPU and memory usage percentage.

\subsubsection{Network Profiler} represents the real-time network information, including the RSSI (Received Signal Strength Indicator) of the Wi-Fi connection from each edge device or nearby router respective to each individual robot. 

The profilers capture the run-time dynamism of the system as the tasks are offloaded or pre-initiated to the computing resources and present the snapshot of the changing environment at each time step. This data is crucial for optimizing task allocation and resource utilization, ensuring seamless performance even under dynamic workloads.

\section{Adaptive Utility Maximization Collaborative Scheduling} 

\subsection{System Model} 
Suppose there are $n$ ground robots R denoted as R = $\{R_{1}, R_{2}, R_{3}, ....R_{n}\}$, and each robot has a computational intensive collaborative task $T$ needing to offload to the edge servers $E$s denoted as $E$ = $\{\epsilon_{1}, \epsilon_{2}, \epsilon_{3}, ....\epsilon_{m}\}$ for processing.  The robot interacts with the edge servers through wireless communication.

\subsection{Problem Formulation}

This paper adopts a utility maximization framework in which each computational resource has an associated utility function, and the objective is to maximize the sum of all utilities at each decision iteration. The basic idea of our algorithm consists of each computation iteration corresponding to the computation of a global collaborative task $T$, we try to find the best computational resource with maximum utility that optimizes the system resources and implements it.

The resource optimization decision-making considers both each resource's computational workload and the servers' proximity to the robot. Thus, we derive the computational and communication model of the system as follows. In robot-robot collaboration, the robot offloads to another robot; however, we denote this robot as an edge server $\epsilon$ for simplification.

The computational model includes processing utility $\eta$ based on CPU for an edge  server $\epsilon$ is quantified by: 

\begin{equation}
\label{eq:cpu_metric}
    \eta_{\epsilon} = \frac{{\gamma_{\epsilon}  - \beta_{\epsilon}}}{{\gamma_{\epsilon}}} 
cpu\end{equation}
where $\gamma_{\epsilon}$ denotes the maximal computation resource (CPU percentage) of the edge $\epsilon$, and $\beta_{\epsilon}$ is the available CPU capacity of edge $\epsilon$. 

Since memory access is tightly coupled with the task execution, we describe the total memory utility of the computation model by $\theta_{\epsilon}$ where $\delta_{\epsilon}$ is the maximum memory limit of the edge device, and $\mu_{i}$ the available memory capacity of the edge $\epsilon$

\begin{equation}
\label{eq:memory_metric}
    \theta_{\epsilon} = \frac{{\delta_{\epsilon} - \theta_i - \mu_{\epsilon}}}{{\delta_{\epsilon}}} 
\end{equation}
The CPU and memory availability is assessed by the performance counters mentioned previously as profilers. 

According to the system model, multiple edge servers are connected as a connected graph with $R_{n}$ and $E_{m}$ over a wireless link or path $f_{mn}$. We denote the RSSI received over $f_{mn}$ from $R_{n}$ to edge server $E_{m}$ as $\lambda_{\epsilon}$. The quality of $f_{mn}$ is an important variable in determining the resource for eventual offloading.

For each $f_{mn}$, we denote its RSSI-based utility as $\kappa_{\epsilon}$ which is a function that calculates the utility of the associated ${f_{mn}}$ between $R_{n}$ and $E_{m}$ based on RSSI. 

\begin{equation}
\label{eq:network_metric}
    \kappa_{\epsilon} = \frac{{\lambda_{\epsilon} - \nu_i}}{{\rho_{\epsilon} - \nu_i }} ,
\end{equation}
where $\lambda_{\epsilon}$ is the current RSSI received by each $\epsilon$ and $\nu_{i}$ the minimum RSSI required for offloading and $\rho_{\epsilon}$ to be the maximum achievable RSSI.

Thus total utility is represented by the $U_{T_\epsilon}$ and is equal to the sum of $\eta$,  $\sigma$, and  $\kappa$, weighted by the variables $\omega_{\eta}$, $\omega_{\sigma}$ and $\omega_{\kappa}$ respectively. We can either allocate different weights to processing or memory utility or provide a single weight to the entire computing model.

\begin{equation}
\label{eq:total_UT_eq}
 U_{T_{\epsilon}} = \omega_{\eta} \cdot \eta_{\epsilon} + \omega_{\sigma} \cdot \sigma_{\epsilon} + \omega_{\kappa} \cdot \kappa_{\epsilon} 
\end{equation}

Here, all the utilities have been normalized to values between 0 and 1, and the sum of the weights equals 1.

Drawing from the concept of the law of diminishing marginal utility, we can establish that sum total utility of each edge server exhibits convex and non-decreasing behavior in the context of resource usage, meaning that as the resource allocation increases, the marginal gain or benefit derived from each additional unit of the resource diminishes. In other words, the utility functions exhibit diminishing returns as more resources are allocated in each iteration.

By combining the utilities proposed from the computation and communication model, we have the total utility formulated as follows:
\begin{equation}
\label{eq:total_UT_formula}
    \mathcal{U}_{T_{\epsilon}}= \sum_{m=1}^E
    ( U_{T_{\epsilon}} )
\end{equation}

Expressed succinctly, the scheduling problem can be given by the following problem:
\begin{equation}
\label{eq:maximization_problem}
    \textrm {max} \: \mathcal{U}_{T_{\epsilon}}^{total}  =
    \sum_{m=1}^E
    (\omega_{\eta} \cdot \eta_{\epsilon} + \omega_{\sigma} \cdot \sigma_{\epsilon} + \omega_{\kappa} \cdot \kappa_{\epsilon}  )
\end{equation}
  \begin{equation}
    \textrm{s.t.} \quad
    \omega_{\eta}, \omega_{\sigma}, \omega_{\kappa} \in [0...1]
    \end{equation}
    \begin{equation}
       \omega_{\eta} + \omega_{\sigma} + \omega_{\kappa}  =1
    \end{equation}

\section{System Implementation} 
Our model includes distributed gateways, schedulers, and executors, all responsible for routing the task node to the best available edge resource based on the highest utility value determined in the previous section. Fig.~\ref{fig:edge_robotics} presents the thematic diagram showing details of different layers. 

\noindent  \textbf{Gateways}
Gateways are computing nodes responsible for analyzing and preprocessing the data collected from system nodes (including remote resources) to accelerate data preprocessing and reduce transmission delay. As the gateways are deployed on the robots, processing and commuting time for rapid service requests is significantly reduced. Subscribing to real-time system parameters like CPU usage, network quality, and memory monitoring, the gateways keep the most updated data about the current state of the system devices to assist in the ultimate scheduling and offloading tasks.

\noindent \textbf{Scheduler}
The data from the Gateways is utilized by the distributed schedulers across all the robots after the profilers and gateways have been set up in accordance with Alg. 1. Upon receiving the job, the scheduler gathers system data, such as the number of robots and computing resources (edges), and initializes the storage for the $edge\_data$ received from the gateways. This information is then used to determine the robot's overall utility, $UT_{self}$, and it also requests the utility data for other robots, $UT_{\epsilon_{r}}$, in the system by subscribing to their respective utility topics.  

Once the utility is calculated for itself and received for the other robots, it compares the utilities assessed by each robot by applying respective weights  $\omega_{\eta}$, $\omega_{\sigma}$, $\omega_{\kappa}$ received from the rosparam server and combines them by summing up utilities with respect to each resource to identify the resource with maximum utility. If the currently assessed resource is the same as was selected in the previous iteration, $selected\_edge$, it will apply a predefined cost to the resource's respective utility. This cost adds to the current edge's utility and ensures that the same edge gets selected in the next iteration to minimize overlapping and switching between nodes. With that, it also ensures a fair distribution of computational resources. After determining the $max_{\epsilon}$ for the task, the scheduler proposes this assignment to the executor for final offloading. The proposed algorithm performed at the scheduler is presented in Alg. 2.

\begin{algorithm}[t]
\label{alg:profiler}
\KwData{List of edges $\epsilon$, List of robots $R$}
\KwResult{Resource Profilers Initialized}

\For{each edge $\epsilon$ in $E$}{
    Initialize Resource Profilers at edge $\epsilon$\;
}

\For{each robot $r$ in $R$}{
    Activate Gateway for robot $r$\;
    \For{each edge $\epsilon$ in $E$}{
       get $\beta_{\epsilon}$, $\mu_{\epsilon}$, $\lambda_{\epsilon}$ for $\epsilon$; \\
       republish $edge\_data$ w.r.t edge $\epsilon$\;
    }
  Initialize Scheduler;
}
\caption{Resource Profilers and Gateways}
\end{algorithm}

\begin{algorithm}[t]
\label{alg:scheduler}
\KwData{List of edges $\epsilon$, List of robots $R$}
\KwResult{Publishes resource with maximum utility}
\caption{Schedulers}
\textbf{Initialize} Data storage for $\epsilon$ in $E$ \\
\For{each $\epsilon$ in $E$}{
    Get $edge\_data$ \;
    }
Get $selected\_edge$ 

\For{each $r$ in $R$}{
    get ${U}_{T_{\epsilon_{r}}}$ \;
}

\SetKwFunction{FUtilityCalc}{Calculate\_Utility}
\FUtilityCalc{}{ \\
    \For{each $\epsilon$ in $E$}{
        Calculate $UT_{self}$ using Eq.~\eqref{eq:cpu_metric},
        Eq.~\eqref{eq:memory_metric}, Eq.~\eqref{eq:network_metric} \\
        Get $\omega_{\eta}$, $\omega_{\sigma}$, $\omega_{\kappa}$ from rosparam server \\
        Apply Eq.~\eqref{eq:total_UT_eq} for $UT_{self}$\\
        \If{$\epsilon$ is $selected\_edge$}{
            APPLY\_COST(get cost) 
            }
        }
\KwRet{$UT_{self}$}
}

\SetKwFunction{FUtilityComp}{Compare\_Utility}
\FUtilityComp{}{\\
    $joint\_ut$ $\gets$ \{** $UT\_{self}$, ** $U_{T_{{\epsilon}_{r}}}$\} \\
    $sums\_\epsilon\_UT \gets \{\}$ \\
    \For{each $\epsilon$ in $\{\epsilon \,|\, robot\_dict \in \text{joint\_ut.values()}\}$}{
        \For{each $\epsilon \in \text{robot\_dict.keys()}$}{
            $sums\_\epsilon\_UT[\epsilon] \gets \sum_{\text{robot\_dict}} \text{robot\_dict}[\epsilon]$ \\
        }
    $max\_e \leftarrow \argmax( \sum (\epsilon\_UT)$ \\
    Publish $max\_\epsilon$ as $selected\_edge$ \\
    }
\KwRet{$max\_\epsilon$} 
}
\end{algorithm}

\noindent \textbf{Executor} 
The edge node assigned to the collaborative task is subsequently transferred to executors running on the robots.

\noindent \textbf{Consensus} The executor maintains an ultimate allocation layer where it receives the decision from all the robot schedulers to perform a final consensus. The computing resource that most of the robots have selected is ultimately scheduled for final offloading. A distributed consensus algorithm \cite{parasuraman2019consensus} is chosen to select the robot that best optimizes the current performance metrics.

\noindent \textbf{Topic Remappings} The executor exploits ROS's nodes launching mechanism to remap all the initially subscribed and published topics to new ones in order to intercept communication to and from the computing resources. Since task re-initialization adds to the handover task latency, topic remappings, on the other hand, allow messages to be exchanged between the computing entities through original and remapped topics. 

\noindent \textbf{Minimizing Reallocation} To reduce offloading to the same computational device, the executor maintains a memory of previous allocations and avoids remapping if the task has been reallocated. This mechanism minimizes unnecessary overhead, leading to faster execution times. 
The proposed algorithm performed at the executor is presented in Alg. 3.

\begin{algorithm}[t]
\caption{Executors}
\label{alg:executor}
\KwData{List of in and out topics $\tau$, List of mapping $\mathcal{M}$}
\KwResult{Offloads data collectively to the selected edge}
\SetKwFunction{FConsensus}{Consensus}
\SetKwFunction{FOffload}{Offload}

\textbf{Initialize} \\
\For{each $s$ in $schedulers$}{
Subscribe for $max_{\epsilon}$ and $selected\_edge$ \\
{Consensus}{($max_{\epsilon}$)}
}

\FConsensus{$max_{\epsilon}$}{\\ 
$S$ $\leftarrow$ \texttt{$s$.values())} \\
\texttt{$e^{*}$} $\leftarrow$ \{($max_{\epsilon}$, \text{count}($max_{\epsilon}$, S)) $\mid$ $max_{\epsilon}$ $\in$ \text{set}(S)\} \\
\If{$e^{*}$ != $selected\_edge$}{
    Offload($e^{*}$) \\
}
Publish $e^*$ \\
\KwRet{$e^{*}$} 
}

\FOffload{$e^{*}$}{ \\
    mapping\_key = $e^{*}$ \;
    \If {mapping\_key $\in \tau$}{ 
        do $remapping[in][out]$ \;
        
        publish $transform[in][out]$ \;
}
\KwRet\;
}
\end{algorithm}

\section {Experiment Setup}
\begin{figure}[t]
    \centering
    \includegraphics[width=0.99\linewidth]{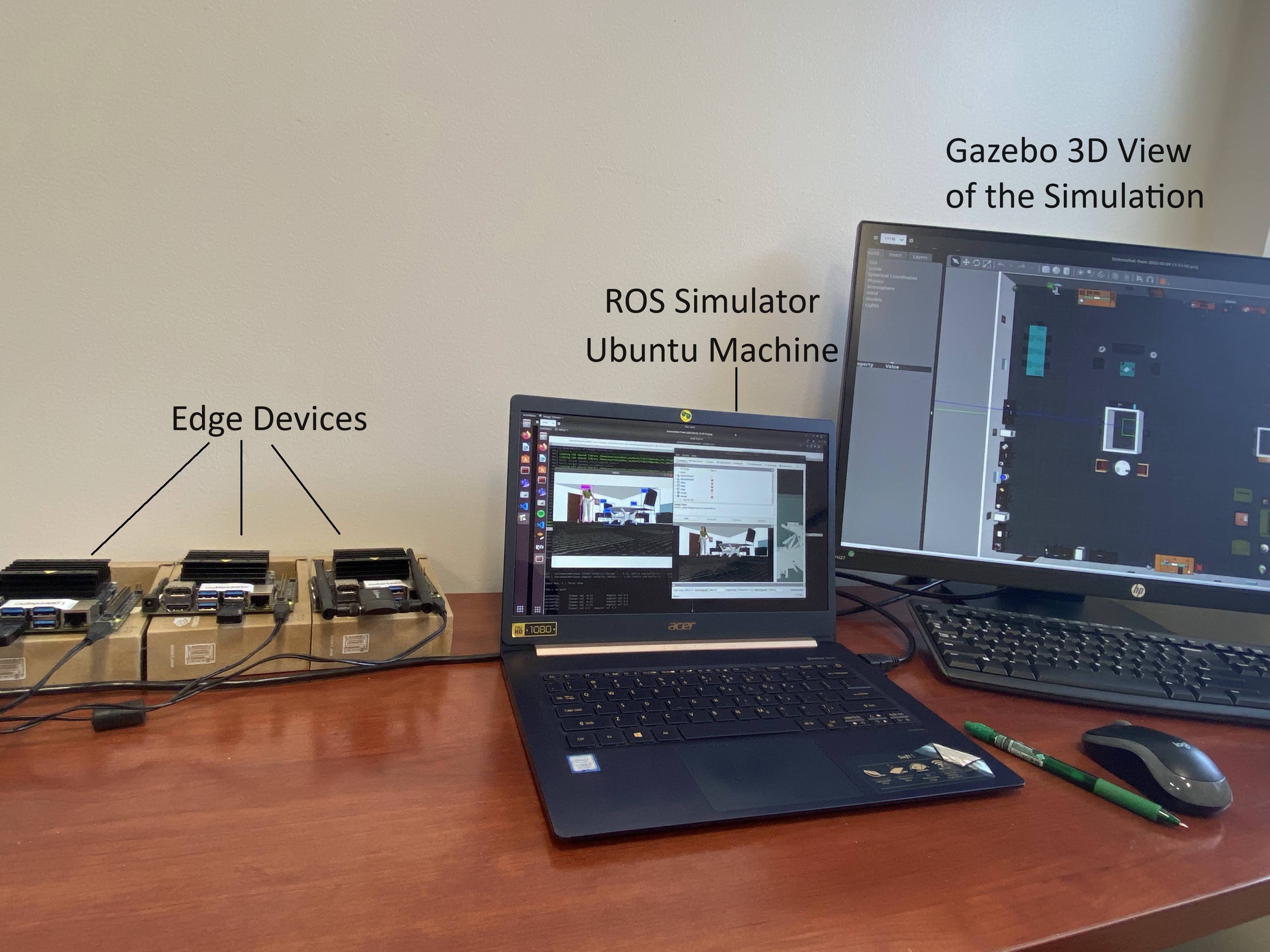}
    \caption{Experiment setup involving one master machine (where three robots are simulated in ROS) and three Edge devices connected via a wireless network.}
    \label{fig:sim_setup}
\end{figure}

We employ a multi-robot simulation environment employing three Turtlebot3 robots and three edge nodes using the ROS-based robotics simulator Gazebo. The simulation system has an Intel Core i7-8565U CPU running at 1.80GHz, 8GB of RAM, and ROS Noetic on a Ubuntu Server. Three edge devices running ARMv8 Processor rev 1(v8l) x 4 architectures, with 4GB RAM and NVIDIA Tegra X1 nvgpu integrated each with Ubuntu 20.04.4 LTS operating system running ROS Noetic, are connected to the multi-robot system through a wireless network. See Fig.~\ref{fig:sim_setup} for an overview of the experiment setup.

\begin{figure*}
  \centering
  \includegraphics[width=.99\textwidth]{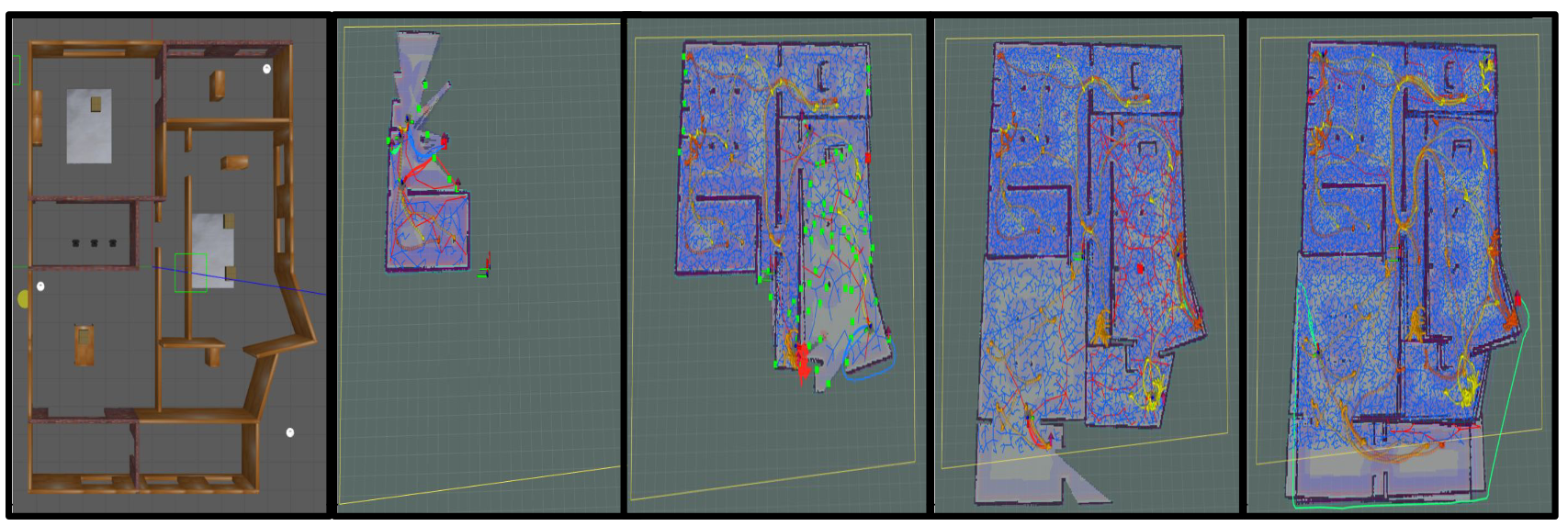} 
  \caption{Simulations in ROS (left): Gazebo world where multi-robots perform collaborative mapping. (right): Incremental performance of map merging process through collaborative random exploration of multi-robots in the simulated space.}
  \label{fig:map_merging_progress}
\end{figure*}

The experimental scenario utilizes mapping performed through Gmapping SLAM, navigation by using ROS navigation, object detection and classification through using YOLOv5 and Kinect depth cameras, and map merging by ROS package multirobot\_map\_merge.

To evaluate our proposed scheduler, we implement a multi-robot cooperative map merging algorithm as a collaborative task for multi-robot cooperative scheduling as shown in Fig.~\ref{fig:map_merging_progress}. The objective for the robots is to decide which edge they have to use to run this map merging algorithm.

The robots are initialized at a ROS master computer and run their respective SLAM modules on the edge servers by fixed predefined allocation. Once the executor offloads the map merge node to the selected edge server, the robots perform autonomous exploration on multiple rapidly exploring randomized (RRT) trees \cite{8202319}. The task is completed once all the robots have collaboratively explored 10\% of the area (which takes around 10 minutes per trial) because of the large size of the environment. Until then, robots will keep exploring space using the ROS navigation stack. A node injects the Yolov5 model randomly on any computing device to allow realistic computing and networking scenarios and to test the performance efficiency under varying traffic loads.

\noindent \textbf{Task Scenarios:} The task is tested in two scenarios: 
\begin{itemize}
\item Scenario 1: involves robots performing individual mapping through SLAM and collaborative map merging by static deployment on the edge servers. 
\item Scenario 2: involves individual SLAM mapping statically initialized at the edge devices while the schedulers and executors take care of collective offloading of collaborative map merging node to the available edge servers for multi-robot task execution. 
\end{itemize}

\noindent \textbf{Comparison schemes:} We compare the performance of three computational offloading schemes in terms of task latency and optimal use of remote resources. 
\begin{itemize}
    \item \textit{Fixed Offloading} The collaborative task is offloaded to the edges statically, and no handover or change occurs in their assignment until the task is fully executed. We implement map merge nodes on edges 1, 2, and 3 for Fixed\_e1, Fixed\_e2, and Fixed\_e3 scenarios, respectively. 
    \item \textit{Dynamic offloading} The task is offloaded to the edges according to the proposed adaptive utility maximization collaborative scheduling. We test the following weight variants of the dynamic offloading scheme: CPU-based, memory-based, and both variants as dyna\_cpu, dyna\_mem, and dyna\_both. We skip the network-based variant for testing in more realistic scenarios.  
\end{itemize}

\noindent \textbf{Evaluation metrics:} To evaluate the performance of different offloading schemes, we analyze CPU utilization, memory utilization, network throughput, task Latency (total execution time of the given task), SSIM (Structural Similarity Index Measurement) (to assess the accuracy of the map merging by comparing with the ground truth) and frequency (Hz) of the $map$ topic.

\begin{figure*}
    \centering
    \begin{subfigure}[b]{0.325\linewidth}
        \includegraphics[width=\linewidth]{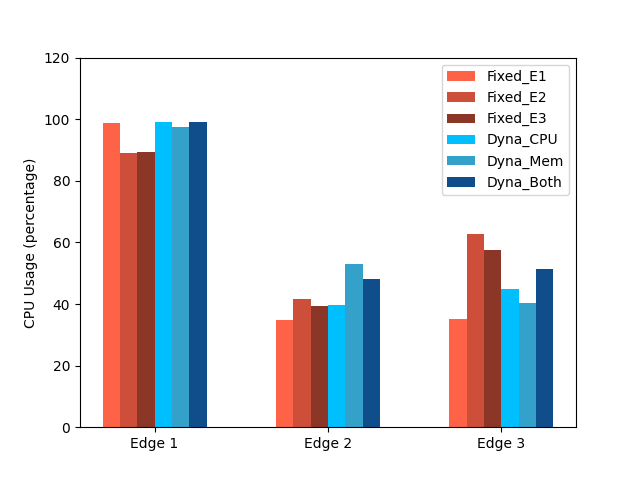}
        \caption{CPU Utilization}
        \label{fig:cpu_graph}
    \end{subfigure}
    \begin{subfigure}[b]{0.325\linewidth}
        \includegraphics[width=\linewidth]{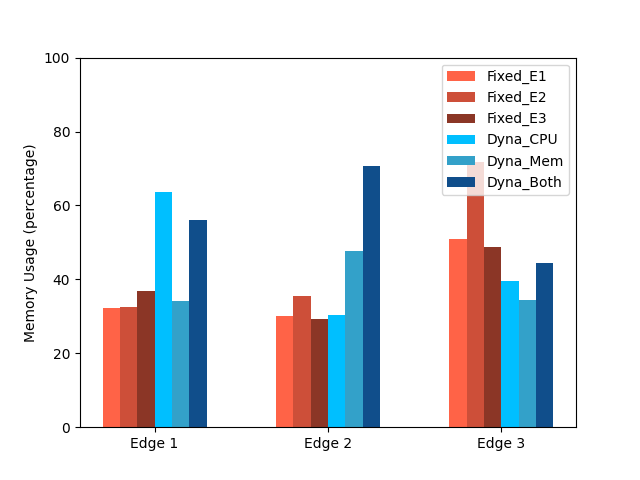}
        \caption{Memory Utilization}
        \label{fig:mem_graph}
    \end{subfigure}
    \begin{subfigure}[b]{0.325\linewidth}
        \includegraphics[width=\linewidth]{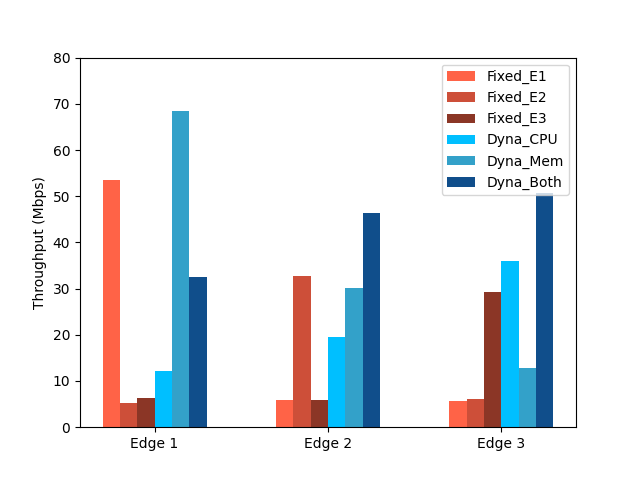}
        \caption{Throughput}
        \label{fig:thpt_graph}
    \end{subfigure}
    
    
    \begin{subfigure}[b]{0.325\linewidth}
        \includegraphics[width=\linewidth]{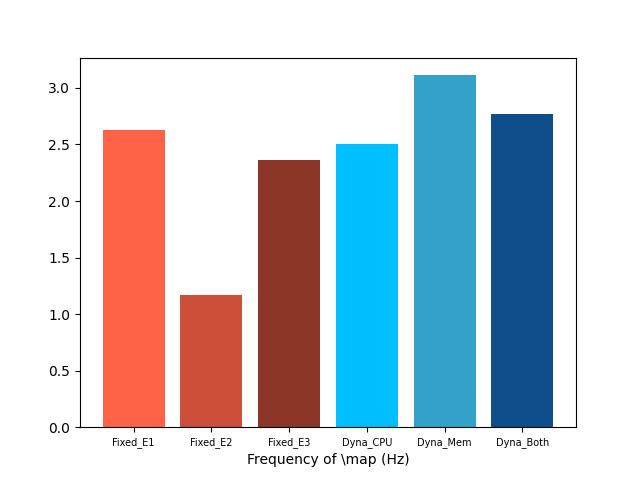}
        \caption{Frequency of Mapping}
        \label{fig:frequency_graph}
    \end{subfigure}
    \begin{subfigure}[b]{0.325\linewidth}
        \includegraphics[width=\linewidth]{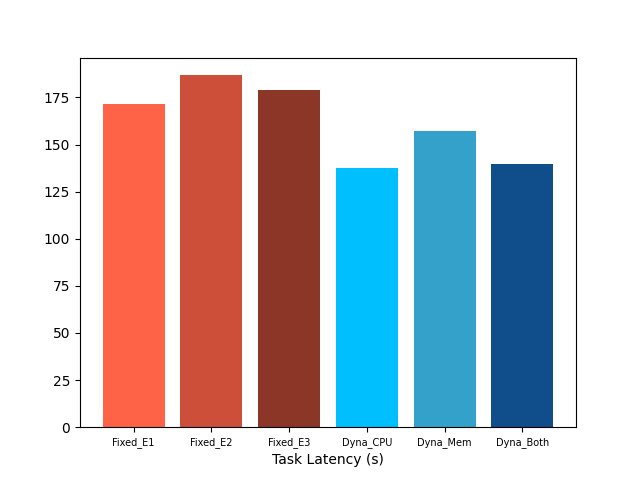}
        \caption{Task Latency}
        \label{fig:latency_graph}
    \end{subfigure}
    \begin{subfigure}[b]{0.325\linewidth}
        \includegraphics[width=\linewidth]{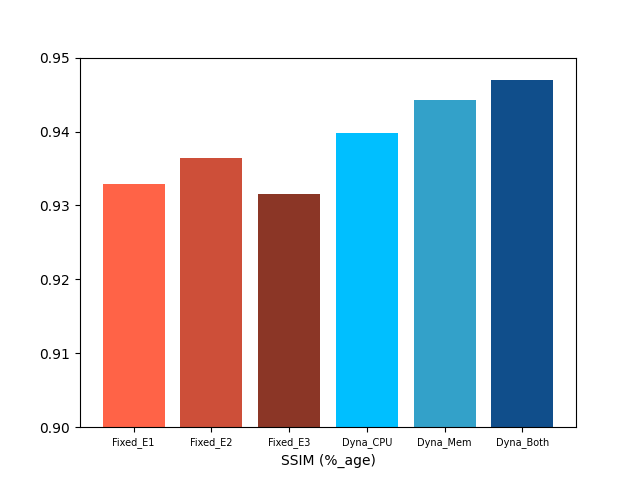}
        \caption{SSIM of Map received}
        \label{fig:ssim_graph}
    \end{subfigure}
    
    \caption{Performance results of evaluation metrics in Fixed and Dynamic Scenarios}
    \label{fig:all_graphs}
\end{figure*}

\section{Performance Evaluation} 
We tested the offloading strategies in the aforementioned scenarios. The results are averaged over 5 independent runs per scenario per strategy. 

\subsection{Effect on Computational Resources}

\textit{CPU Utilization:} It is the peak CPU percentage usage when the task is initialized on the computing servers. Fig. ~\ref{fig:cpu_graph} shows the effect of offloading the collaborative task on the different edge servers. Here, six offloading schemes are compared for resource optimization assessment. The proposed scheme evaluated under dynamic offloading fared better in even workload across all available resources. Due to the heterogeneity in the CPU cores, we observe Edge 1 having higher usage relative to its peers. This is because Edge 1 has fewer CPU cores than the others. Due to SLAM node initialization and random injection of Yolov5 models, it is also observed to be working at its maximal CPU capacity.

\textit{Memory Utilization:} Similarly, in Fig.~\ref{fig:mem_graph}, the average memory utilization's memory \% (mean value) of the edge servers varies significantly compared to the fixed variants. The proposed scheme tried to balance the memory overload due to the random projection of computational loads by ensuring none of the servers exhausted their maximum memory capacity. The data shows that the proposed scheme handled the memory resources optimally with an apparent curve based on the most current changing system conditions. We see an anomaly in Edge 3's behavior for fixed\_e2 case, which was due to pre-congestion of its memory even before the task execution (initial memory usage: 31.8\%, with Yolov5 injection: 87\%-92.1\% during the trial).

\textit{Network Throughput:} We observe increased throughput (40\% $\uparrow$) by better management of computational and networking resources through the proposed dynamic offloading scheme compared to the fixed variants as shown in Fig.~\ref{fig:thpt_graph}. In fixed variants, only the edges executing the task were seen to perform at maximum throughput; however, in the dynamic variants, due to task hopping between available resources, the throughput is seen changing at different resources due to their respective network quality. 

\subsection{Effect on Task Performance}
\textit{Frequency of Mapping} Since the proposed strategy performs optimal resource allocation, the edge devices performing map merging receive a better frequency of overall $map$ topic than the fixed variants with improved throughput and better performance through resource balancing. The dyna\_mem variant with more weight allocation to the memory module in the computational model outperforms the rest of the dynamic variants by 20\% increase in frequency refer to Fig.~\ref{fig:frequency_graph}.

\textit{Task Latency} The task latency improved (~ 28\%) across all variants compared to the fixed allocation schemes as indicated by Fig~\ref{fig:latency_graph}. Since the utility-aware offloading strategy uses multiple distant resources as a pool to execute various tasks in tandem simultaneously, this contributes to shorter job execution times and improved system performance, thereby meeting the enhanced performance objective.

\textit{Map Accuracy} As shown in Fig.~\ref{fig:ssim_graph}, the map accuracy improves for the robots randomly exploring the environment through the collaborative task through all utility-aware dynamic offloading variants with dyna\_both achieving maximum accuracy compared to the ground truth. 


\begin{figure}[ht]
    \centering
    \includegraphics[width=\linewidth]{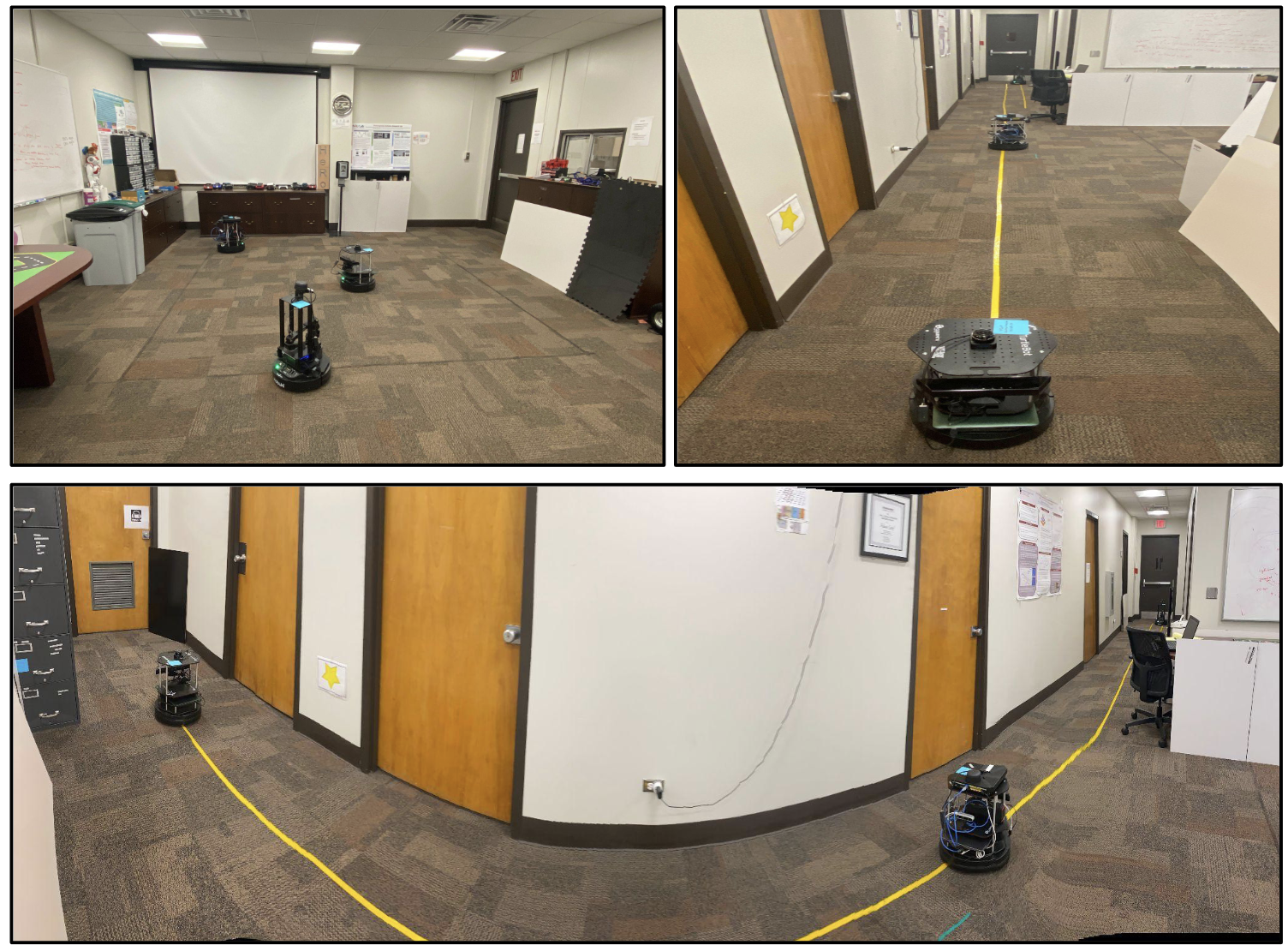}
    \caption{Experiment setup involving three robots connected via a wireless network performing multi-robot map merging.}
    \label{fig:phy_setup}
\end{figure}

\begin{figure}[ht]
    \centering
    \includegraphics[width=.99\linewidth]{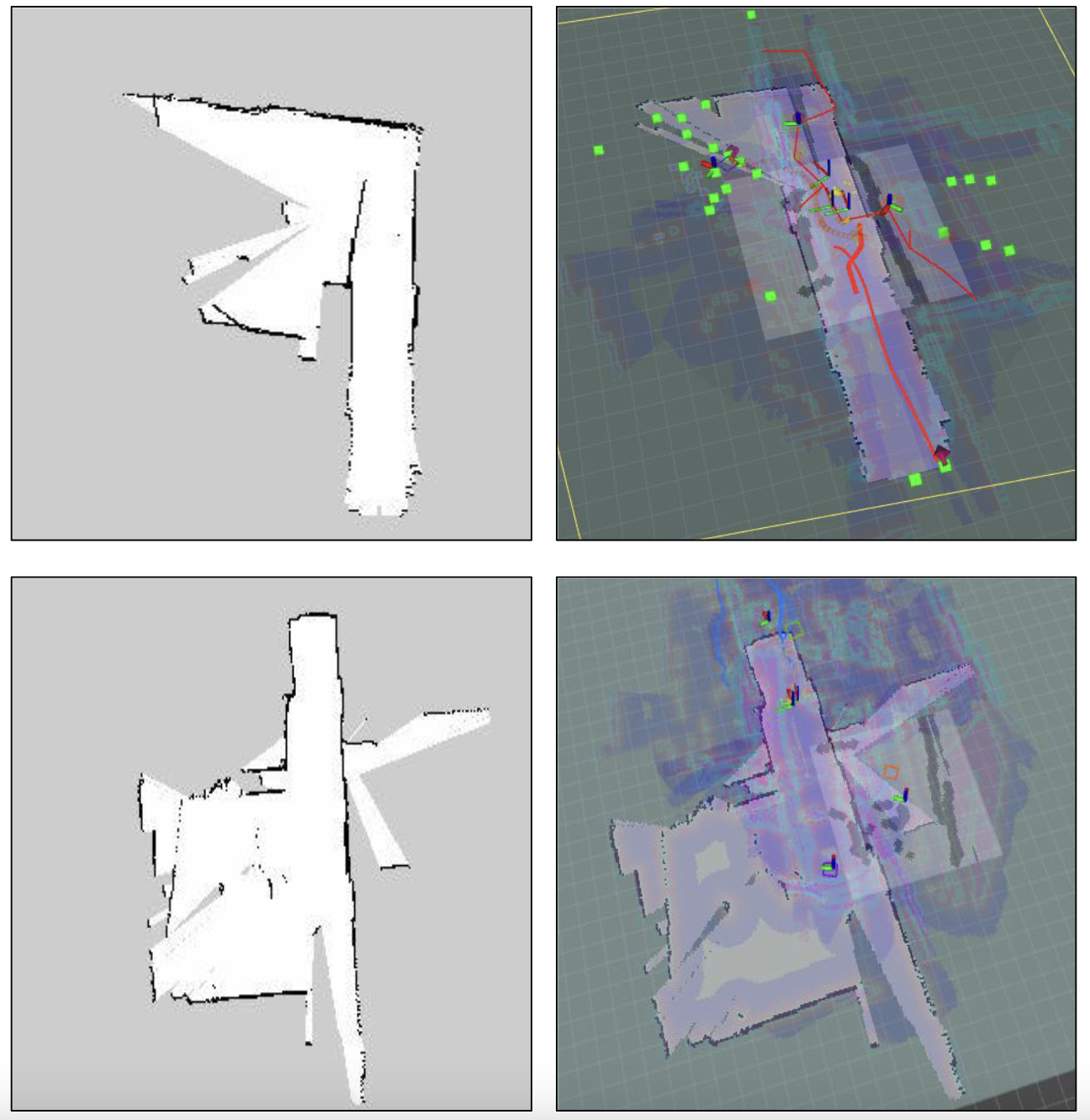}
    \caption{Maps obtained through Fixed offloading (\textit{top}) and Dynamic offloading (\textit{bottom}) after 10\% of map completion.}
    \label{fig:evaluation_phy_exp}
    \vspace{-2mm}
\end{figure}

\begin{figure}[ht]
    \centering
    \includegraphics[width=.54\linewidth]{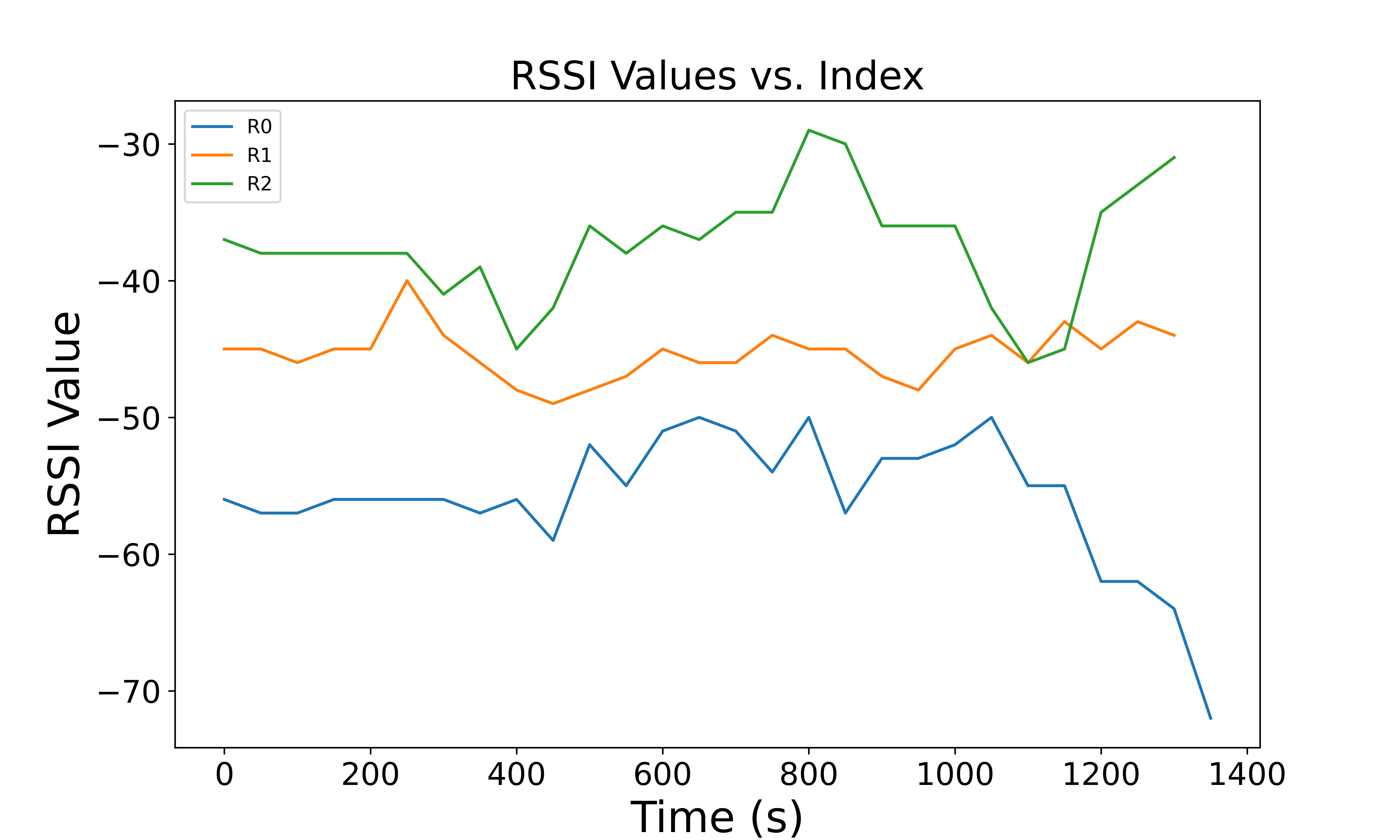}
    \includegraphics[width=.44\linewidth]{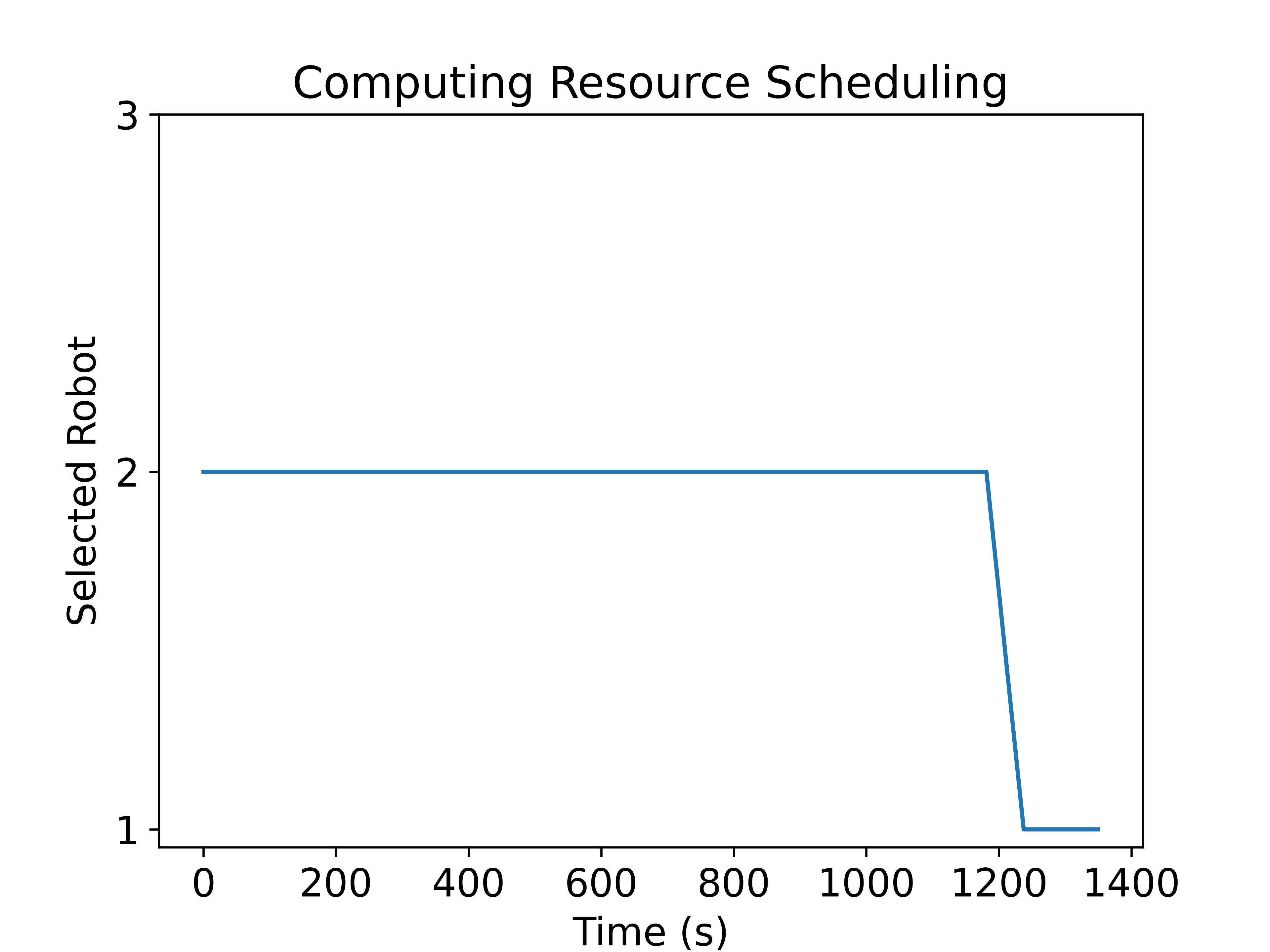}
    \caption{(RSSI receptions on different robots throughout the experiment (\textit{left}). Robots switch tasks between each other as the signal reception changes throughout the trial (\textit{right}).}
    \label{fig:rssi_switch_phy_exp}
    \vspace{-2mm}
\end{figure}

\subsection{Real world demonstration} 
A real-world implementation was performed to validate the efficacy of the proposed strategy by tasking multi-robot collaborative map-merging to a cohort of three Turtlebot 2e robots (Fig.~\ref{fig:phy_setup}) that collectively leveraged each other's computational resources for the task (no edge devices were deployed). The experimentation focused on the network variant only due to the impracticality of simulating realistic network conditions.
The comparison schemes akin to those conducted in simulated environments indicate that dynamic offloading based on the proposed strategy outperforms fixed offloading in terms of map frequency and accuracy, as indicated in Fig.~\ref{fig:evaluation_phy_exp}. 
The results of the experimental demonstration indicate that the proposed strategy consistently selected the robot with the highest Received Signal Strength Indicator (RSSI) at each iteration and offloaded the map-merging task to the robot with optimal signal reception, as noticed by a small segment of the experiments presented in Fig.~\ref{fig:rssi_switch_phy_exp}. For instance, when the network conditions degraded (monitored through the RSSI metric) for robot 2, the scheduling of the map merging task was moved to robot 1 through a consensus algorithm to optimize the cooperative task performance.

\section{Conclusion}
This work presents and evaluates a unique task offloading approach for latency-sensitive edge-based multi-robot systems. The approach proposes a utility-maximization-based task offloading mechanism to minimize total service time and maximize resource consumption by profiling devices (such as CPU utilization, network conditions, and memory needs), as well as performing dynamic computational offloading decision mechanisms. We compare our proposed approach with the static offloading strategy to analyze task performances of computing entities for map merging and random autonomous exploration.
Our future work would focus on integrating the network module of the communication model with realistic network conditions in real-world scenarios and robots. We hope to test both robot-robot and robot-edge-robot collaboration schemes as proposed through this work.

\bibliography{references}
\bibliographystyle{IEEEtran}
\end{document}